\documentclass[letterpaper,10pt,conference]{ieeeconf}

\IEEEoverridecommandlockouts 
\PassOptionsToPackage{table}{xcolor} 
\usepackage{amsmath} 
\usepackage{amssymb} 
\usepackage{amsfonts}
\usepackage[sort,compress]{cite}
\usepackage[usenames,dvipsnames]{xcolor}
\usepackage{graphicx}
\usepackage{tabularx}
\usepackage{multirow}
\usepackage{mathtools}
\usepackage{esvect} 
\usepackage[]{siunitx}
\usepackage{caption}
\usepackage[nolist]{acronym}
\usepackage[pdftex,pdfstartview={FitV},pdfpagelayout={TwoColumnLeft},bookmarksopen=true,plainpages=false,colorlinks=true,linkcolor=black,citecolor=black,urlcolor=black,filecolor=black ,pagebackref=false,hypertexnames=false,plainpages=false,pdfpagelabels]{hyperref}
\usepackage[T1]{fontenc} 
\usepackage{mathtools, cuted} 
\usepackage{bm}
\usepackage{xspace}
\usepackage{arydshln}

\usepackage{etoolbox}
\usepackage[normalem]{ulem}
\usepackage{empheq}
\usepackage{cleveref}
\robustify\bfseries
\robustify\uline

\newcolumntype{H}{>{\setbox0=\hbox\bgroup}c<{\egroup}@{}}

\usepackage{enumerate}
\usepackage{balance} 
\usepackage{booktabs} 

\newcommand{\NoRed}{\textcolor{red}{No}}
\newcommand{\YesGreen}{\textcolor{ForestGreen}{Yes}}

\usepackage{makecell}

\usepackage{amsmath,calc}

\usepackage{tikz} 
\usetikzlibrary{shapes,snakes} 
\usetikzlibrary{tikzmark} 
\usetikzlibrary{calc, arrows.meta, positioning} 
\usetikzlibrary{shapes.multipart} 
\usepackage{subcaption} 

\usepackage{footnote}
\makesavenoteenv{tabular}
\makesavenoteenv{table}

\ifdefined\qtyproduct
\else
  \ifdefined\NewCommandCopy
    \NewCommandCopy\qtyproduct\SI
  \else
    \NewDocumentCommand\qtyproduct{O{}mm}{\SI[#1]{#2}{#3}}
  \fi
\fi

\DeclareCaptionLabelSeparator{periodspace}{.\quad}
\definecolor{opt_color}{RGB}{203,255,182}
\definecolor{check_color}{RGB}{195,218,255}
\definecolor{recheck_color}{RGB}{255,176,176}
\definecolor{delaycheck_color}{RGB}{255,228,181}

\definecolor{agent1_color}{RGB}{234,153,153}
\definecolor{agent2_color}{RGB}{151,104,175}
\definecolor{agent3_color}{RGB}{249,203,156}
\definecolor{agent4_color}{RGB}{194,115,156}
\definecolor{agent5_color}{RGB}{159,197,232}
\definecolor{agent6_color}{RGB}{117,186,117}
\definecolor{obstacle_color}{RGB}{160,117,109}

\usepackage{mdframed}

\usepackage{algorithm}
\usepackage{algpseudocode}

\usepackage{mathtools,amssymb,lipsum}
\newacro{CGD}{Constraint-Guided Diffusion}
\newacro{slam}[SLAM]{Simultaneous Localization and Mapping}
\newacro{uav}[UAV]{Unmanned Aerial Vehicle}
\acrodefplural{uav}[UAVs]{unmanned aerial vehicles}
\newacro{gns}[GNS]{Global Navigation Satellite}
\newacro{gnss}[GNSS]{Global Navigation Satellite System}
\acrodefplural{gnss}[GNSS]{Global Navigation Satellite Systems}
\newacro{mcl}[MCL]{Monte-Carlo localization}
\newacro{imu}[IMU]{Inertial Measurement Unit}
\newacro{dof}[DOF]{degree-of-freedom}
\acrodefplural{dof}[DOFs]{degrees-of-freedom}
\newacro{ransac}[RANSAC]{random sample consensus}
\newacro{map}[MAP]{maximum a posteriori}
\newacro{mle}[MLE]{maximum likelihood estimation}
\newacro{rms}[RMS]{root-mean-square}
\newacro{dem}[DEM]{digital elevation model}
\newacro{vio}[VIO]{visual-inertial odometry}
\newacro{cnn}[CNN]{convolutional neural network}
\newacro{pdf}[pdf]{probability density function}
\newacro{ahrs}[AHRS]{attitude and heading reference system}
\newacro{lidar}[LIDAR]{light detection and ranging}
\newacro{relu}[ReLU]{rectified linear unit}
\newacro{rtk}[RTK]{real-time kinematic}
\newacro{gps}[GPS]{global positioning system}
\newacro{fcn}[FCN]{fully-connected network}
\newacro{brm}[BRM]{building ratio map}
\newacro{sfm}[SfM]{Structure-from-Motion}
\newacro{vpr}[VPR]{visual place recognition}
\newacro{fov}[FOV]{field of view}
\newacro{poc}[POC]{partially overlapping circular}

\newcommand{\ourmethod}{CGD} 

\title{\LARGE \bf \ourmethod: Constraint-Guided Diffusion Policies \\ for UAV Trajectory Planning}

\author{Kota Kondo, Andrea Tagliabue$^*$, Xiaoyi Cai$^*$, Claudius Tewari, Olivia Garcia, \\ Marcos Espitia-Alvarez, and Jonathan P.\ How
	\thanks{The authors are with the Department of Aeronautics and Astronautics, Massachusetts Institute of Technology. *Equal contribution.
	    {\texttt{\{kkondo, atagliab, xyc, cttewari, ogarcia, mfea, jhow\}@mit.edu.}}}
    \thanks{This work is funded by the Air Force Office of Scientific Research MURI FA9550-19-1-0386.
    The authors thank Rachel Sun for her insightful comments and feedback.}
}%

\begin{document}

\thispagestyle{plain}
\pagestyle{plain}

\maketitle
\begin{abstract} 
Traditional optimization-based planners, while effective, suffer from high computational costs, resulting in slow trajectory generation. A successful strategy to reduce computation time involves using Imitation Learning (IL) to develop fast neural network (NN) policies from those planners, which are treated as expert demonstrators. Although the resulting NN policies are effective at quickly generating trajectories similar to those from the expert, (1) their output does not explicitly account for dynamic feasibility, and (2) the policies do not accommodate changes in the constraints different from those used during training.

To overcome these limitations, we propose Constraint-Guided Diffusion (\ourmethod{}), a novel IL-based approach to trajectory planning. \ourmethod{} leverages a hybrid learning/online optimization scheme that combines diffusion policies with a surrogate efficient optimization problem, enabling the generation of collision-free, dynamically feasible trajectories. The key ideas of \ourmethod{} include dividing the original challenging optimization problem solved by the expert into two more manageable sub-problems: (a) efficiently finding collision-free paths, and (b) determining a dynamically-feasible time-parametrization for those paths to obtain a trajectory. Compared to conventional neural network architectures, we demonstrate through numerical evaluations significant improvements in performance and dynamic feasibility under scenarios with new constraints never encountered during training.
\end{abstract}

\section{Introduction}\label{sec:introduction}


Traditional constrained optimization-based approaches for obstacle avoidance and trajectory planning, while achieving impressive performance \cite{tordesillas2020mader, tordesillas2021panther, zhou2021raptor, bry2011rapidly, allen2016real}, have been limited in their onboard deployment by their large computational cost. 
A promising approach to reduce computation time is to leverage Imitation Learning (IL)-based strategies~\cite{tordesillas2023deep, Tagliabue2021DemonstrationEfficientGP, loquercio2021learning, kaufmann2020deep, kahn2017plato, zhang2016learning}, where a fast neural network (NN) policy is trained offline to imitate the diverse demonstrations collected by computationally expensive optimization-based planners, significantly reducing the online computational costs. However, (1) \textit{preserving the rich distribution of trajectories produced by the expert}, while (2) \textit{guaranteeing dynamic feasibility and collision avoidance} in the NNs' output remain two open challenges.

\begin{figure}[!t]
    \centering
    \includegraphics[width=\columnwidth, trim={8cm 6cm 8cm 6cm}, clip]{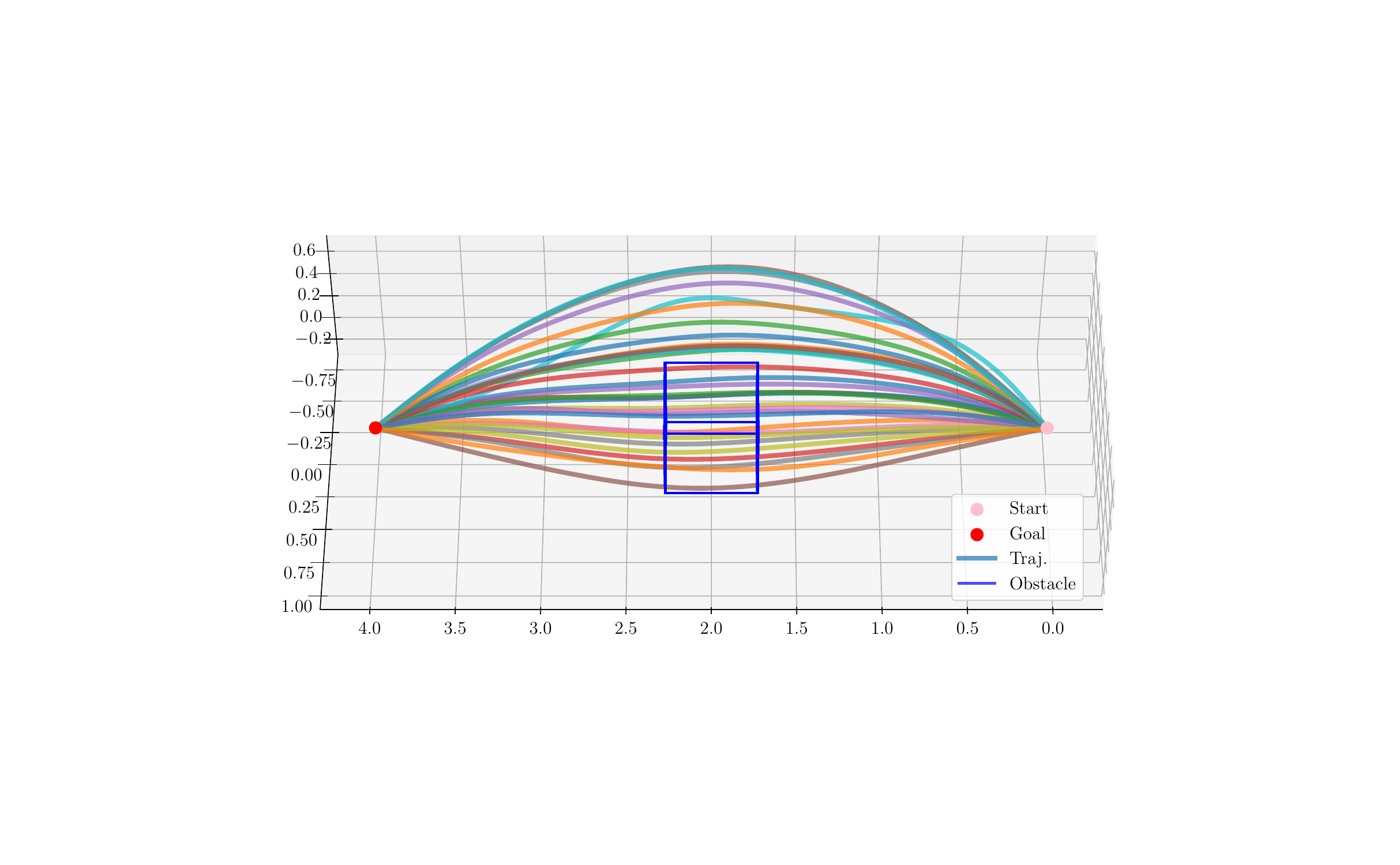}
    \includegraphics[width=\columnwidth, trim={8.5cm 7cm 8.5cm 9cm}, clip]{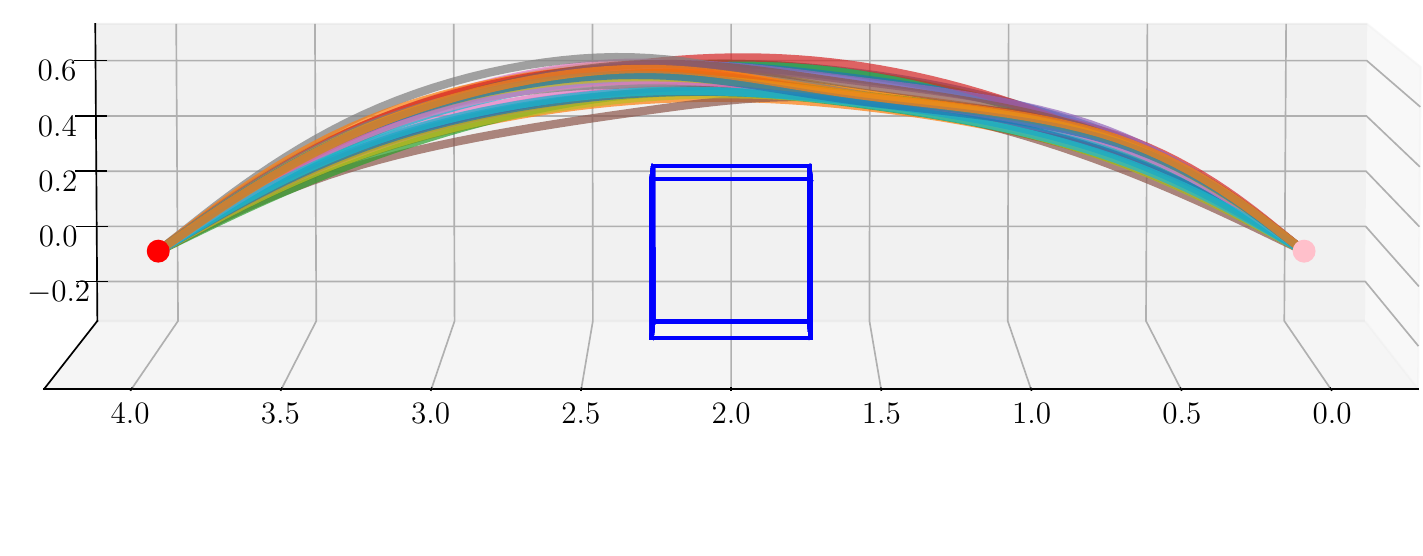}
    \caption{Constraint-Guided Diffusion (CGD) is a method that efficiently generates collision-free and dynamically feasible trajectories. This figure shows 32 trajectories generated by CGD, capturing multi-modality continuously.\label{fig:multimodality}}
    \vspace{-2em}
\end{figure}

Recently, diffusion models~\cite{ho2020denoising}, a type of NN that can iteratively generate arbitrary samples from a learned training data distribution, have shown impressive performance in learning highly-multi-modal data distributions with applications in areas such as computer vision and natural language processing (see survey in~\cite{yang2023diffusion}). Because diffusion models capture the multi-modality of the training dataset, as shown in Fig.~\ref{fig:multimodality}, they are better alternatives to multi-layer perceptrons (MLPs) for learning navigation trajectories. This is because there are often multiple feasible paths to take, and naively applying IL to a navigation dataset may lead to mode collapse, resulting in sub-optimal and even unsafe trajectories~\cite{tordesillas2023deep}. On the other hand, diffusion models can be trained without specially designed multi-modal losses, making them easy to train and attractive for robotic tasks~\cite{janner2022planning, xiao2023safediffuser, sridhar2023nomad}. 
However, ensuring constraint satisfaction in trajectories sampled from diffusion models remains a fundamental open question for safe real-world deployments.

A strategy commonly employed to ensure constraint satisfaction in NNs is to modify the nominal output via a surrogate optimization problem. This problem is either (1) smaller than the expert's, enforcing a subset of key constraints such as dynamic feasibility, acting as a safety filter (e.g.,\cite{xiao2023safediffuser, christopher2024projected}), or (2) the same as the expert's, leveraging the output of the policy as an initial guess (e.g.,\cite{li2024efficient, giannone2024aligning}). However, if the \textit{constraints imposed at deployment differ from those used to train the policy}, it (1) could fail, owing to the fact that it only optimizes a subset of the original variables, and (2) may require a long time to converge.

To address these challenges, this paper proposes Constraint-Guided Diffusion (\ourmethod{}), a novel IL-based approach to trajectory planning. \ourmethod{} leverages diffusion policies and a surrogate optimization problem that can be solved efficiently, enabling the generation of collision-free, dynamically feasible trajectories.

Key ideas of \ourmethod{} involve dividing the original challenging optimization problem into two more tractable sub-problems that can be solved iteratively, inspired by projected gradient descent. The first sub-problem focuses on training an efficient diffusion policy that outputs high-performance, multi-modal trajectories. This provides a high-quality initial guess for the subsequent surrogate optimization problem, which aims to enforce constraint satisfaction in every iteration performed by the diffusion model. More specifically, the surrogate optimization problem targets two coupled constraints: (1) collision avoidance and (2) dynamic feasibility. Inspired by block coordinate descent, we address constraint (a) by modifying the intermediate trajectories generated by the diffusion model based on their gradients with respect to constraint violation, and tackle constraint (b) by iteratively adjusting the time parametrization of the trajectory and (c) by solving a Quadratic Program (QP) that ensures dynamic feasibility.

\subsection{Contributions:} 
\begin{itemize}
\item A novel block-coordinate-descent-inspired method to modify the intermediate trajectories from diffusion to encourage constraint satisfaction.
\item CGD can handle time-varying constraints or constraints at deployment time that differ from ones imposed at training time. This enables significant performance improvements compared to traditional MLP-based policies under constraints that differ from the ones used during training.
\item By evaluating our framework in simulations on the challenging task of obstacle avoidance for an agile UAV, we show that CGD enables computational vs performance trade-offs, unlike previous approaches that are not iterative and have fixed computational cost and performance.
\end{itemize}

\section{RELATED WORKS}
\subsection{Learning with Guaranteed Constraint Satisfaction}
A promising method to achieve constraint satisfaction in neural networks (NN) involves designing ad-hoc network architectures or training procedures that inherently ensure their outputs satisfy the required constraints. DC3~\cite{donti2021dc3} is one of the first works in this domain, focusing on time-invariant equality and inequality constraints.

Frerix et al.\cite{frerix2020homogeneous} introduced an approach to train NNs that guarantees the satisfaction of homogeneous linear inequality constraints ($\mathbf{A}\mathbf{x} \leq \mathbf{0}$), and RAYEN\cite{tordesillas2023rayen} proposed a method for enforcing the satisfaction of convex constraints.
While these approaches enable extremely fast computation by transferring the computational burden of enforcing constraints to the offline training phase, they operate under the assumption that the constraints are time-invariant.
This assumption is incompatible with our work, which deals with time-varying constraints due to the dynamic nature of the planning horizon.

Among the works exploring implicit layer-based NN architectures for guaranteed constraint satisfaction, OptNet~\cite{brandon2017optnet} facilitates the imposition of time-varying equality and inequality constraints on the outputs of NNs. 
Although effective, this approach has been primarily investigated in scenarios with convex constraints, whereas our method addresses more general non-convex constraints.

\subsection{Diffusion Model with Constraint Satisfaction}
While diffusion models have demonstrated promising capabilities in learning optimal trajectory distributions for robotic tasks, the trajectories sampled from these models are not guaranteed to satisfy constraints. Recent approaches to enforcing constraints can generally be categorized into two groups:
The first category requires a \textit{post-processing} step, where the diffusion model outputs are used as initial guesses for constrained optimization algorithms in order to speed up convergence (e.g.,~\cite{power2023sampling, li2024efficient, giannone2024aligning}). 
For instance, Li et al.~\cite{li2024efficient} augmented the nominal diffusion training loss with constraint violation terms to encourage the diffusion model to output constraint-satisfying trajectories, which are used to warm-start a non-convex numerical solver. 
In addition, Power et al.~\cite{power2023sampling} encoded constraints as input features to the diffusion policy and used the diffusion outputs as initial guesses for a sampling-based solver.
In comparison, the second category of methods provides additional guidance for the diffusion policy by running \textit{constrained optimization within the denoising loop} to improve the generalization performance of the diffusion policy in out-of-distribution scenarios (e.g.,~\cite{christopher2024projected, xiao2023safediffuser}). 
For example, Christopher et al.~\cite{christopher2024projected} used a non-convex solver to ensure collision-free denoised trajectories. 
Alternatively, Xiao et al.~\cite{xiao2023safediffuser} leveraged a quadratic program (QP) to more efficiently ensure safety by encoding collision avoidance constraints via control barrier functions (CBFs~\cite{ames2019control, xiao2019control}).
While our work also solves a constrained optimization within the denoising loop similar to~\cite{christopher2024projected,xiao2023safediffuser}, the proposed non-convex surrogate optimization problem requires significant computation to solve for planning onboard a UAV.
Therefore, to make the problem tractable, we leverage a block coordinate descent-inspired approach to efficiently solve simpler sub-problems in order to encourage constraint satisfaction.
\begin{figure*}[t]
\centering
\includegraphics[width=\linewidth, trim={0cm 0cm 0cm 0cm},clip,]{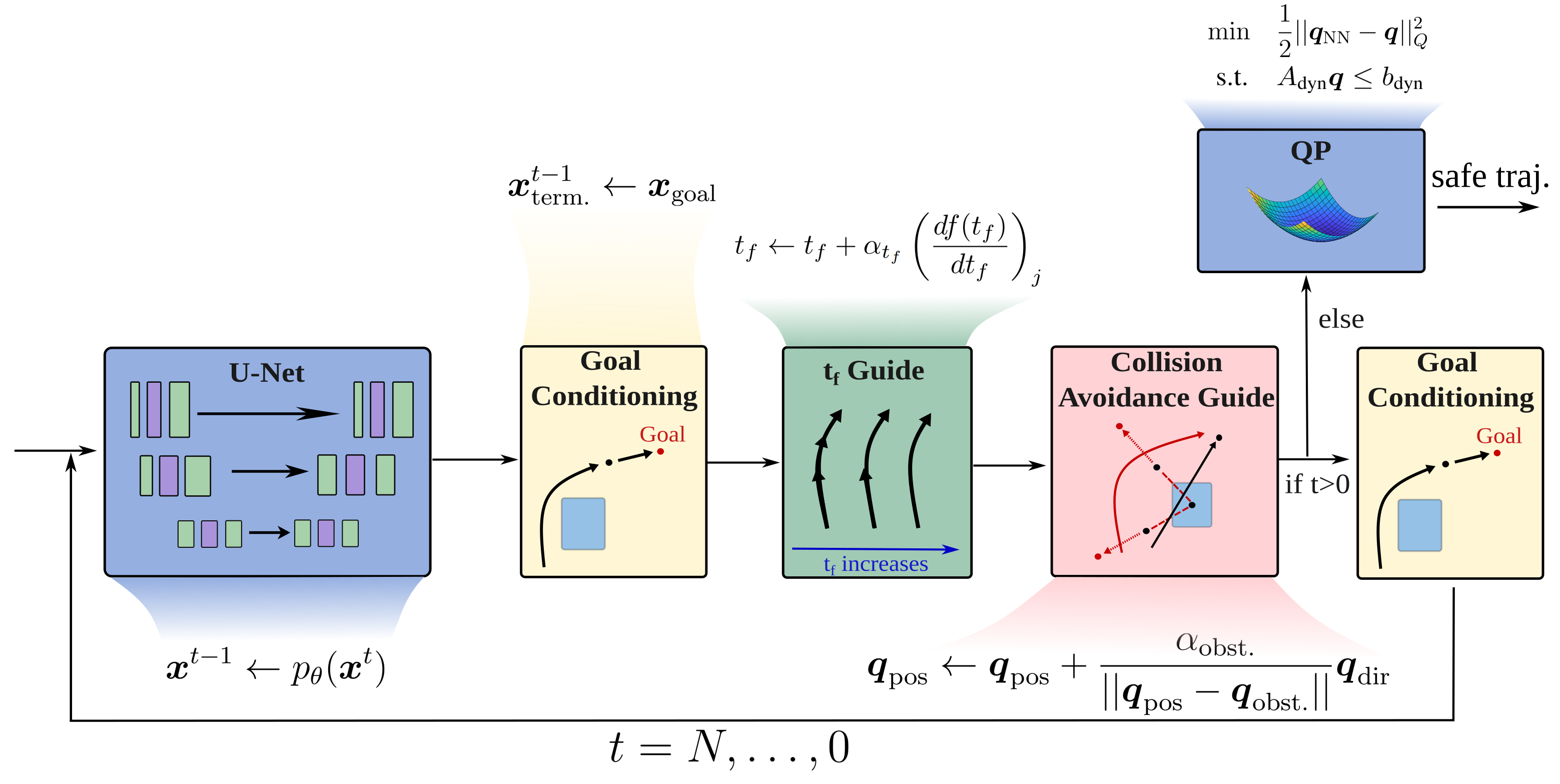}
\caption{
Overview of the proposed approach: 
We employ U-Net for our framework, which is trained to output a trajectory $\boldsymbol{x}^t$ similar to the ones provided by the model-based optimization-based trajectory planner (the expert). 
A standard diffusion model (e.g., \cite{ho2020denoising}) works by iteratively refining (denoising) $\boldsymbol{x}^t$ across the iterations $t=N, \dots, 0$. 
Our work modifies this iterative scheme, introducing multiple modules. 
First, we introduce a goal conditioning block, which forces the generated trajectory to have a terminal state at the desired goal state. 
Second, the time parametrization of the trajectory (described by the total time $t_f$) is adjusted to accommodate for new constraints or imperfect trajectories from the diffusion model (e.g., $t_f$ needs to be increased if the maximum flight speed is decreased, or if the initially generate trajectory cannot reach the goal, as shown in the diagram)). 
Third, the trajectory is modified to ensure collision avoidance constraints. 
Specifically, the trajectory is altered based on the distance of the control points from the center of the obstacle $\boldsymbol{q}_{\text{obst.}}$.
Last, a Quadratic Program (QP) is solved to ensure that the trajectories satisfy dynamic feasibility constraints. 
Note that the QP cannot directly optimize the total time $t_f$, hence justifying the presence of the $t_f$-Guide block. 
Note that the $N$ iterations procedure should be performed at every planning timestep.
}
\vspace{-2em}
\label{fig:architecture}
\end{figure*}

\section{Approach}\label{sec:approach}

\subsection{Overview}
Fig. \ref{fig:architecture} provides an overview of the approach, which is conceptually divided in the following parts. 
First, the task of finding trajectories for obstacle avoidance is addressed by leveraging a diffusion policy (we use a U-Net), trained via IL from expert demonstrations. 
This part represents the most computationally challenging aspects of the original optimization problem, as obstacle avoidance typically requires solving hard non-convex optimization problems. 
In this context, diffusion policies not only enable rapid inference speed but also capture the highly multi-modal trajectory distributions typically encountered in obstacle avoidance and provide the ability to impose further terminal state constraints (Goal Conditioning), as demonstrated in \cite{janner2022planning}. 

Then, the more tractable problem of finding a dynamically feasible time parametrization for the previously generated trajectory is solved online via an efficient surrogate optimization problem.
Specifically, we formulate a block-coordinate descent-inspired optimization procedure that alternates between (a) finding a new time parametrization for the trajectory (changing the total time $t_f$), (b) shifting control points for collision avoidance, and (c) enforcing dynamic feasibility of the control points by solving a Quadratic Program (QP). This procedure is  solved in conjunction with the denoising iterations performed by the diffusion model, guiding the iterative denoising steps and enabling performance and computational trade-offs.



\subsection{Diffusion Model Architecture}

\subsubsection{Expert}

To perform imitation learning, we use PANTHER*~\cite{tordesillas2023deep, tordesillas2021panther} as an expert, which is an optimization-based trajectory planner that generates a trajectory that satisfies collision avoidance and dynamic constraints such as velocity, acceleration, and jerk, while tracking an obstacle.
PANTHER* generates a trajectory parameterized by clamped uniform B-splines and MINVO basis~\cite{tordesillas2022minvo}, which are defined by position control points (detailed in MADER~\cite{tordesillas2020mader} and Appendix).
The weights in the cost functions are defined as: $\alpha_{\boldsymbol{j}}=10^{-6}$, $\alpha_{\psi}=10^{-6}$, $\alpha_{\text{T}}=8.0$, $\alpha_{\boldsymbol{g}_{\boldsymbol{p}}}=10^{3}$, $\alpha_{g_{\psi}}=0.0$, and $\alpha_{\text{FOV}}=10^{2}$.
Refer to \cite{tordesillas2023deep, tordesillas2021panther} for the notation.

\subsubsection{Student NNs architectures}

This section provides an overview of the student NNs architectures.
Diffusion models are a class of generative models that through a series of denoising steps, generate noise-free data.
\begin{equation}
    \boldsymbol{x}_{t-1} \sim p_{\theta}(\boldsymbol{x}_{t-1} | \boldsymbol{x}_{t})
\end{equation}
where $\boldsymbol{x}_{t-1}$ is the input data, $\boldsymbol{x}_{t}$ is the output (denoised) data, and $p_{\theta}$ is the diffusion model, and $\theta$ is the set of parameters of the NNs.

More precisely, NNs are trained to predict noise that is added to the input data, and then the predicted noise is used to denoise the input data.
For the NNs, we employ U-Net~\cite{ronneberger2015unet} as the architecture for diffusion models.
We denote this noise predector as $\boldsymbol{\epsilon}_{\theta}$.
We implemented DDPM~\cite{ho2020denoising}, where $p_{\theta}$ is defined as:
\begin{equation}
    p_{\theta}(\boldsymbol{x}_{t-1} | \boldsymbol{x}_{t}) = \frac{1}{\sqrt{\alpha_{t}}} \left( \boldsymbol{x}_{t} - \frac{\beta_{t}}{\sqrt{1-\bar{\alpha}_{t}}} \boldsymbol{\epsilon}_{\theta}(\boldsymbol{x}_{t}) + \sigma_{t}\boldsymbol{z} \right)
\end{equation}
where $t$ is the denoising steps $t = N, \dots, 1$, $N$ is the total denoising step, $\beta_{t}$ is a scheduled variance of the noise, $\alpha_{t} \vcentcolon= 1 - \beta_{t}$, and $\bar{\alpha}_{t} \vcentcolon= \sum_{i=1}^{t} \alpha_{i}$, $\sigma_{t}^2 = \beta_{t}$, and $\boldsymbol{z} \sim \mathcal{N}(0, \boldsymbol{I})$.

For the input to the NNs, we use FiLM  conditioning~\cite{perez2017film, chi2024diffusion}.
The inputs consist of and the output is a set of position control points, yaw control points, and the trajectory total time.

For benchmarking in Section~\ref{sec:simultation-results}, we also train Deep-PANTHER-based Multilayer perception (MLP) student.
Note that Deep-PANTHER~\cite{tordesillas2023deep} only generates position control points and trajectory total time, and uses a closed-form solution for yaw, since it has an explicit tracking term. (See PUMA \cite{kondo2023puma} for explicit and implicit tracking).
However, other approaches such as PUMA~\cite{kondo2023puma} optimize position as well as yaw control points, and considering our approaches' generalizability to other methods, we include yaw control points in action variables.

\subsection{Surrogate Optimization Modules}

While the original optimization-based PANTHER* approach~\cite{tordesillas2021panther, tordesillas2023deep} optimizes its cost function by using control points and the total trajectory time as variables, it is non-convex and computationally intensive. Therefore, we divide the problem into four smaller sub-problems: (1) optimizing the control points for dynamic feasibility, (2) guiding the total trajectory time, (3) directing the control points for collision avoidance, and (4) conditioning the terminal state.

\subsubsection{\textbf{QP Optimization}}

While trained to imitate expert demonstrations, the student neural networks (NNs) do not inherently guarantee constraint satisfaction. Therefore, a module that optimizes the control points for dynamic feasibility is essential to ensure dynamic feasibility.
The optimization problem is formulated as a minimum-deviation quadratic program (QP):
\begin{equation}\label{eq:qp-ctrl-points}
    %
    \begin{aligned}
    \underset{\makecell{\boldsymbol{q}}}{\min} & \quad \frac{1}{2} || \boldsymbol{q_{\text{NN}}} - \boldsymbol{q}||^{2}_{Q} \\
    %
    \text{s.t.} & \quad A_{\text{dyn}} \boldsymbol{q} \leq b_{\text{dyn}} \\
    \end{aligned}
\end{equation}
where $\boldsymbol{q_{\text{NN}}}$ is control points generated by the student NNs, $\boldsymbol{q}$ is the control points to be optimized, $Q$ is a positive definite matrix, and $A_{\text{dyn}}$ and $b_{\text{dyn}}$ are the dynamic constraints on the control points (detailed in Appendix). 
The dynamic constraints depend on the trajectory total time, $t_f$, and therefore are time-varying. 
The purpose of this QP is to generate control points that satisfy the dynamic constraints while being close to the control points generated by the student NNs.

\subsubsection{\textbf{$t_f$ Guide}}
While the above QP module optimizes the control points for dynamic feasibility, it does not change the trajectory total time.
Therefore, we need a module that guides the trajectory total time to generate a trajectory that satisfies the dynamic constraints.
The reason why we do not solve the optimization problem to change the trajectory total time, $t_f$ is that $t_f$ is that dynamic constraint defined by $A_{\text{dyn}}$ and $b_{\text{dyn}}$ becomes nonlinear when optimizing $t_f$, and therefore, it requires additional overhead to impose dynamic constraints in the trajectory total time optimization.
To guide $t_f$, we propose an innovative constraint gradient-based guide.
This step is inspired by \cite{janner2022planning}, where they guide the trajectory based on the reward function in diffusion denoising steps.
The difference from \cite{janner2022planning} is that we guide the trajectory total time based on the gradient of the dynamic constraints, not a reward function.

Since the dynamic constraints are defined as $A_{\text{dyn}} \boldsymbol{q} \leq b_{\text{dyn}}$, by defining $f(t_f) = b_{\text{dyn}} - A_{\text{dyn}} \boldsymbol{q}$, the gradient of the dynamic constraints with respect to $t_f$ is:
\begin{equation}
    \frac{df(t_f)}{dt_f} = -\frac{dA_{\text{dyn}}}{dt_f} \boldsymbol{q}
\end{equation}
Note that the minimum element of $f(t_f)$ corresponds to the largest violation of the dynamic constraints.
Therefore, by defining $i$ as the index of the minimum element of $f(t_f)$, we can guide $t_f$ as follows: 
\begin{equation}\label{eq:tf-guide}
    t_f \leftarrow t_f + \alpha_{t_f} \left(\frac{df(t_f)}{dt_f}\right)_{j}
\end{equation}
where $\alpha_{t_f}$ is a step size, and $(\cdot)_{j}$ is the $j$th element of the vector.


\subsubsection{\textbf{Goal Conditioning}}

As demonstraited in~\cite{janner2022planning}, diffusion models are capable of generating samples seamlessly under the given constraints.
We leverage this property to condition the terminal state of the trajectory.
Specifically, we replace the last position control points to be the goal position as suggested in~\cite{janner2022planning}.

\subsubsection{\textbf{Collision avoidance guide}}

The original PANTHER*~\cite{tordesillas2021panther, tordesillas2023deep} solves non-convex optimization problems to generate a trajectory that satisfies collision avoidance constraints.
However, the non-convex optimization problem is computationally expensive, and therefore we push control points away from obstacles using a simple collision avoidance guide.
For each control point, we calculate the direction from the center of an obstacle to the point, $\boldsymbol{q}_{\text{dir}} = \frac{\boldsymbol{q}_{\text{pos}} - \boldsymbol{q}_{\text{obst.}}}{||\boldsymbol{q}_{\text{pos}} - \boldsymbol{q}_{\text{obst.}}||}$, $\boldsymbol{q}_{\text{pos}}$ is each control point, and $\boldsymbol{q}_{\text{obst.}}$ is the center of the obstacle to avoid. 
We then push each control point away from the obstacle as follows:
\begin{equation}\label{eq:collision-avoidance-guide}
    \begin{aligned}
    \boldsymbol{q}_{\text{pos}} & \leftarrow \boldsymbol{q}_{\text{pos}} + \frac{\alpha_{\text{obst.}}}{||\boldsymbol{q}_{\text{pos}} - \boldsymbol{q}_{\text{obst.}}||} \boldsymbol{q}_{\text{dir}} \
    \end{aligned}
\end{equation}
where $\alpha_{\text{obst.}}$ is a constant pushing scale, and in our case set to 1.0.
The reason why we have the distance in the denominator of the pushing scale is that we want to push the control points away from the obstacle when the control points are close to the obstacle, and we want to push the control points less when the control points are far from the obstacle.

\subsection{Diffusion Model-based Constraint Satisfied Trajectory Generation Framework}

The previous sections has described the diffusion model's architecture and the surrogate optimization modules.
This section provides an overview of the diffusion model-based trajectory generation framework.
As Algorithm~\ref{alg:diffusion-model-based-safe-trajectory-planner} shows, the framework consists of the following steps:

\begin{enumerate}
    \item This algorithm requires Diffusion model $\mu_{\theta}$, Whilte noise $\boldsymbol{x}^{N}$, and Goal position $\boldsymbol{x}_{\text{goal}}$.
    \item The diffusion model generates position and yaw control points and the trajectory total time. We denote a set of these action variables as $\boldsymbol{x}^{t}$, where the superscript is the denoising time step. As discussed, this output does not guarantee constraint satisfaction.
    \item To reduce terminal goal error, we replace the last position control points, $\boldsymbol{x}^{t-1}_\text{term.}$, with the goal position, $\boldsymbol{x}_{\text{goal}}$. Note that during the subsequent surrogate optimization/guiding steps, the last position control points will be changed to satisfy the dynamic constraints.
    \item The trajectory total time is guided by the total time guiding module based on Eq.~\ref{eq:tf-guide}.
    \item We then apply collision avoidance guide based on Eq.~\ref{eq:collision-avoidance-guide}.
    \item Before the next denoising step, the last position control points are replaced again by the goal position, and this will facilitate the diffusion model to generate a trajectory that ends close to the goal position. Note that this step will be omitted if the last denoising step is reached since imposing the last position control points might violate the dynamic constraints. 
    \item The aforementioned process is repeated until the last denoising step.
    \item Lastly, both the position and yaw control points are optimized in the QP module to satisfy the dynamic constraints, generating trajectory $\boldsymbol{x}$.
\end{enumerate}

\begin{algorithm}
    \caption{Diffusion model-based safe trajectory planner}\label{alg:diffusion-model-based-safe-trajectory-planner}
    \begin{algorithmic}
    \Require $\mu_{\theta}$, $\boldsymbol{x}^{N}$, $\boldsymbol{x}_{\text{goal}}$
    \For{$t = N$, \dots, 0} \ \small{\color{gray}{ // Denoising step}}
        \State $\boldsymbol{x}^{t-1} \leftarrow p_{\theta}(\boldsymbol{x}^{t})$ \ \ \ \ \ \small{\color{gray}{ // Diffusion model and denoising step}}
        \State $\boldsymbol{x}^{t-1}_\text{term.} \leftarrow \boldsymbol{x}_{\text{goal}}$ \quad \ \ \ \ \ \small{\color{gray}{ // Goal conditioning}}
        \State $t_f \leftarrow t_f + \alpha_{t_f} (\frac{df(t_f)}{dt_f})_{j}$ \quad \ \ \ \small{\color{gray}{// $t_f$ guide}}
        \State $\boldsymbol{q}_{\text{pos}} \leftarrow \boldsymbol{q}_{\text{pos}} + \frac{\alpha_{\text{obst.}}}{||\boldsymbol{q}_{\text{pos}} - \boldsymbol{q}_{\text{obst.}}||} \boldsymbol{q}_{\text{dir}}$ \small{\color{gray}{// Collision avoidance guide}}
        \If{$t > 0$}
        \State $\boldsymbol{x}^{t-1}_\text{term.} \leftarrow \boldsymbol{x}_{\text{goal}}$ \small{\color{gray}{// Goal conditioning}}
        \EndIf
    \EndFor
    \State $\boldsymbol{x} \leftarrow \text{QP}(\boldsymbol{x}^{0})$ \quad \small{\color{gray}{// QP optimization}}
    \end{algorithmic}
\end{algorithm}

\subsection{Training the Diffusion Model}

This section outlines the training process for the diffusion model. The model generates a set of position and yaw control points along with the total time of the trajectory.

During the training of the diffusion model, we employ the Mean Squared Error (MSE) loss. 
It is important to note that for the Multi-Layer Perceptron (MLP), which serves as a benchmark in our study, we utilize the assignment loss proposed by Deep-PANTHER~\cite{tordesillas2023deep}. 
This distinction is crucial because, as highlighted in Section~\ref{sec:introduction}, the diffusion model excels at capturing multimodal distributions, in contrast to the MLP, which is prone to mode collapse.
\section{Simulation Results}\label{sec:simultation-results}

\subsection{In-distribution (ID) Benchmarking}

To showcase our approach's advantage, we benchmark our approach to the optimization-based expert and Deep-PANTHER-based MLP.
We first showcase how the QP ensures constraints are satisfied.
Then we show how our surrogate optimization modules (goal conditioning, $t_f$ guiding, collision avoidance, and QP) improve the performance of the students.
We also perform an ablation study, where we do not use some of the surrogate optimization modules. 

All the simulation results are obtained with the same hyperparameters, and the results are the average of the best trajectory out of \textit{\# of rollouts} trajectories generated for each simulation over 100 simulations.
Note that the expert's trajectory generation process is parallelized for \textit{\# of rollouts}.
The simulations were performed on a desktop CPU with Intel\textregistered \ Core\texttrademark \ i9-9900K processors and 64GB of RAM.

\subsubsection{Effects of QP}

\begin{figure}
    \centering
    \subcaptionbox{In-distribution MLP performance.\label{fig:id-benchmark-mlp}}{
    \begin{tikzpicture}[every text node part/.style={align=center}]
    \node {\includegraphics[width=\columnwidth, trim={8cm 6cm 8cm 6cm}, clip]{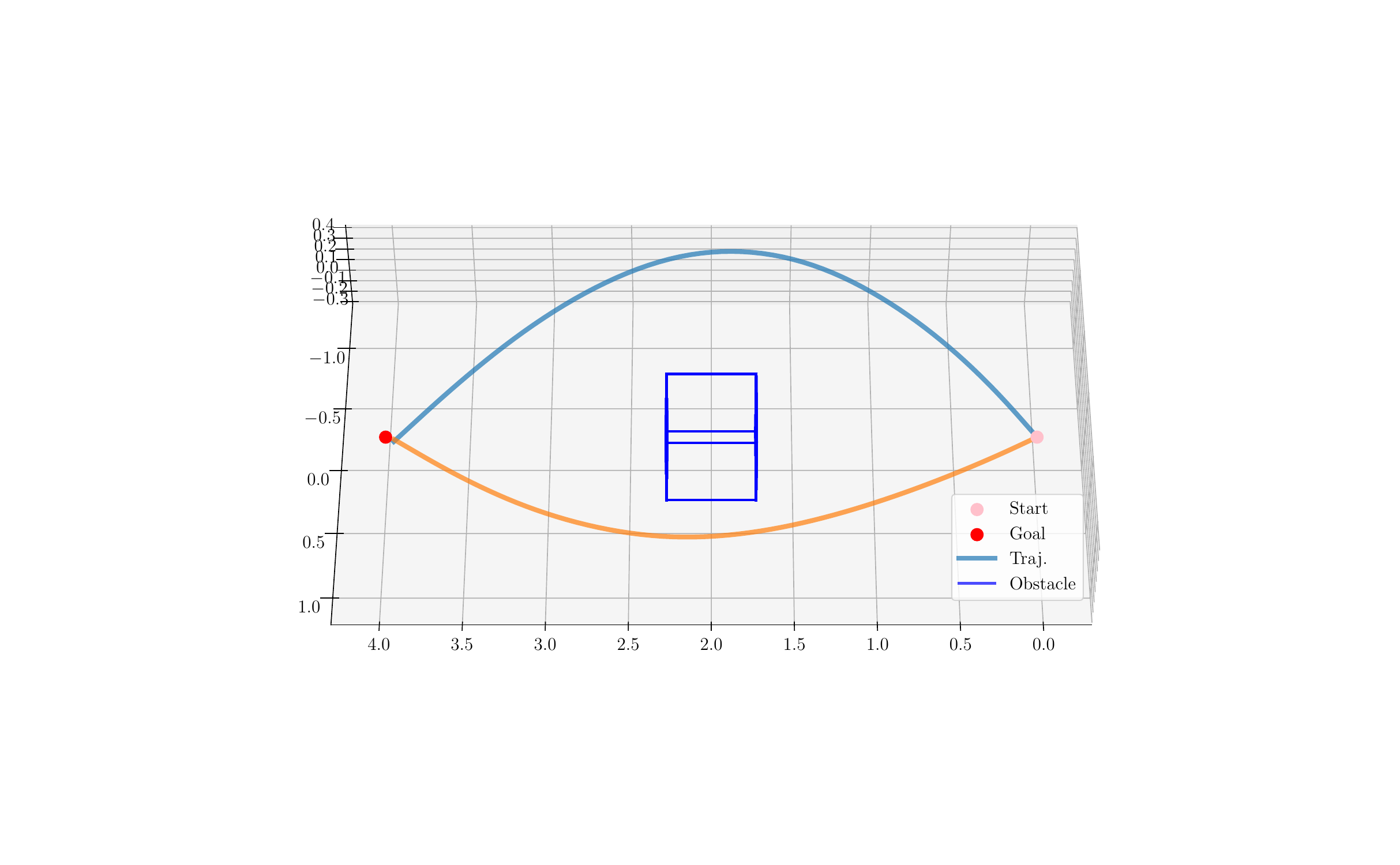}};
    \end{tikzpicture}}
    \subcaptionbox{In-distribution DDPM performance.\label{fig:id-benchmark-ddpm}}{
    \begin{tikzpicture}[every text node part/.style={align=center}]
    \node {\includegraphics[width=\columnwidth, trim={8cm 6cm 8cm 6cm}, clip]{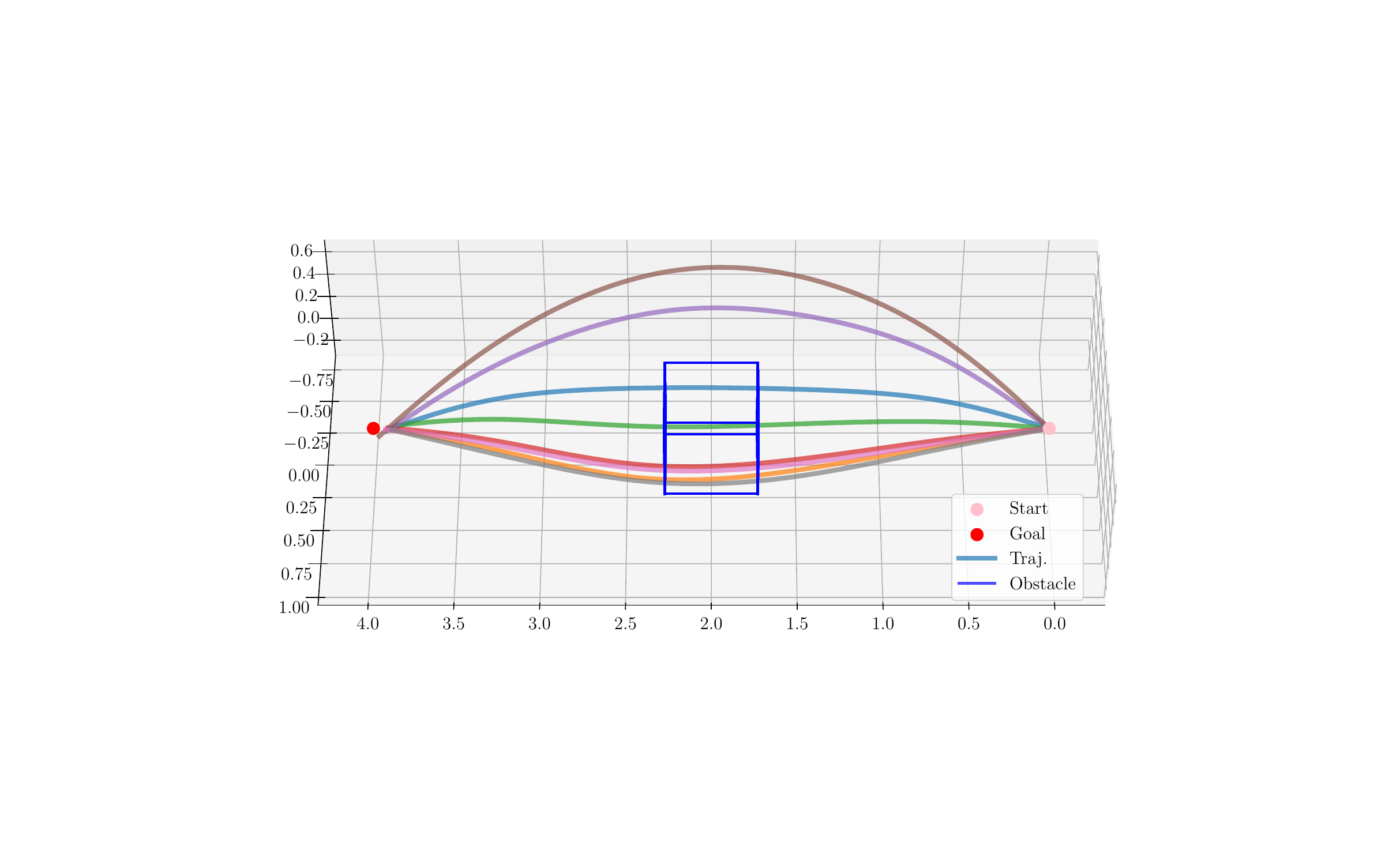}};
    \end{tikzpicture}}
    \caption{In the context of in-distribution benchmarking, both MLP and DDPM meet the constraints and generates low-cost trajectories as anticipated. 
    However, DDPM excels in capturing multi-modality, unlike MLP, which in this specific scenario, only identifies two modes. 
    As elaborated in~\cite{tordesillas2023deep}, MLP assigns trajectories to specific modes, and the loss calculation is based on this assignment.
    Trajectories not assigned to any mode do not influence the loss, resulting in no updates to the neural network's weights. 
    Consequently, even though MLP generates a certain number of rollouts, only a subset, specifically two modes in this case, results in qualitatively valuable trajectories. 
    This leaves MLP with six unassigned and, from a computational standpoint, costly untrained trajectories. (We therefore only visualize two trajectories.) 
    On the other hand, all of DDPM's rollouts (eight in this scenario) are effectively utilized, capturing diverse modes.}
    \label{fig:id-benchmark}
    \vspace{-1em}
\end{figure}

We first showcase the effect of the QP.
While without QP the students can generate trajectories that violate dynamic constraints, the QP ensures that the generated trajectories are within the dynamic constraints.
Table~\ref{tab:sim-study_rollout_effects_with_qp} shows with QP, both MLP and Diffusion models achive even better performance than expert while maintaining \SI{0.0}{\%} dynamic constraint violation.
As expected, as the number of rollouts increases, computation time also increases.
However, the performance improvement is not as significant as expected in this scenario.
Fig.~\ref{fig:id-benchmark} demonstrates the superior mode-capturing capability of DDPM compared to MLP. 
As discussed in details in the caption of ~\ref{fig:id-benchmark}, while MLP only captures two modes, DDPM successfully captures various modes in this collision avoidance problem.

\begin{table}[htbp]
    \renewcommand{\arraystretch}{1}
    \setlength{\tabcolsep}{1pt}
    \normalsize
    \centering
    \caption{\centering Study Rollout Effect with QP \textemdash 100 simulations}
    \label{tab:sim-study_rollout_effects_with_qp}
    \resizebox{\columnwidth}{!}{
        \begin{tabular}{>{\centering\arraybackslash}m{0.05\textwidth} >{\centering\arraybackslash}m{0.08\textwidth} >{\centering\arraybackslash}m{0.08\textwidth} >{\centering\arraybackslash}m{0.06\textwidth} >{\centering\arraybackslash}m{0.06\textwidth} >{\centering\arraybackslash}m{0.04\textwidth} >{\centering\arraybackslash}m{0.12\textwidth} >{\centering\arraybackslash}m{0.08\textwidth} >{\centering\arraybackslash}m{0.08\textwidth} >{\centering\arraybackslash}m{0.06\textwidth} }
            \toprule
            & \multicolumn{5}{c}{\textbf{Framework}} & \multicolumn{1}{c}{\textbf{Performance}} & \multicolumn{2}{c}{\textbf{Safety}} & \multirow{2}[10]{*}{\makecell{\textbf{Comp.} \\ \textbf{[ms]}}} \\
            \cmidrule(lr){2-6}
            \cmidrule(lr){7-7}
            \cmidrule(lr){8-9}
            & \# of rollouts & Term. condi. & Colls. avoids. & Guided $t_f$ & QP & Cost & Colls. [\%] & Dyn. violation [\%] & \\
            \midrule
            Expert & 8 & - & - & - & - & 69.5 (17.8) \footnote{In one scenario, the expert's trajectory became stuck in a local minimum, failing to go around the obstacle. This resulted in a trajectory with a significantly higher cost of 5188.0, leading to an average cost of 69.5. Excluding this outlier, the average cost is reduced to 17.8} & 0.0 & 0.0 & 4543 \\
            \midrule
            MLP & 8 & \NoRed{} & \NoRed{} & \NoRed{} & \YesGreen{} & 23.6 & 0.0 & 0.0 & 44.0 \\
            \midrule
            DDPM & 8 & \NoRed{} & \NoRed{} & \NoRed{} & \YesGreen{} & \textcolor{ForestGreen}{\textbf{23.1}} & 0.0 & 0.0 & 91.4 \\
            \bottomrule
        \end{tabular}}
\end{table}

\subsubsection{Generalization to Out-of-distribution (OOD) Scenarios}

To further test the generalization capability of our approach, we consider an OOD deployment scenarios where the dynamic constraints enforced by the QP are tighter than the ones used for training.
The expert demonstrations, with which the students are trained, are generated with $v_{\text{max}}=\SI{2.5}{\m/\s}$, $a_{\text{max}}=\SI{5.5}{\m/\s^2}$, $j_{\text{max}}=\SI{30.0}{\m/\s^3}$, and $\dot{\psi}_{\text{max}}=\SI{5.0}{\deg/\s}$, where $v_{\text{max}}$, $a_{\text{max}}$, $j_{\text{max}}$, and $\dot{\psi}_{\text{max}}$ are the maximum velocity, acceleration, jerk, and yaw rate, respectively.
We test the students on the scenarios where the dynamic constraints are $v_{\text{max}}=\SI{1.25}{\m/\s}$, $a_{\text{max}}=\SI{2.75}{\m/\s^2}$, $j_{\text{max}}=\SI{15.0}{\m/\s^3}$, and $\dot{\psi}_{\text{max}}=\SI{2.5}{\deg/\s}$.
In other words, the dynamic constraints are twice as tight as the training scenarios.
Table~\ref{tab:sim-ood_50_constr} shows the benchmarking results for the OOD scenario with tighter dynamic constraints on the static obstacle. 
First, we tested both MLP and DDPM without any additional constraints, and the results show that both MLP and DDPM violate the dynamic constraints.
Then we tested both MLP and DDPM with QP, and the results show that both MLP and DDPM do not violate the dynamic constraints, but the performance is significantly degraded and we observe a high collision rate.
To mitigate the performance degradation, we tested both MLP and DDPM with QP and guided $t_f$, and the results show that DDPM has much better performance than MLP, but we still observer a small collision rate. 
To further improve the performance, we tested both MLP and DDPM with QP, guided $t_f$, and terminal goal conditioning, and the results show that both MLP and DDPM have much better performance.
Finally to reduce the collision rate, we tested both MLP and DDPM with QP, guided $t_f$, terminal goal conditioning, and collision avoidance, and the results show that both MLP and DDPM achieve \SI{0.0}{\%} collision rate.
With all the modules activated, our proposed approach achieves the smallest costs among the costs achieved by MLP and DDPM that satisfy both collision avoidance and dynamic feasibility constraints).
Note that the computation time is increased as we add more modules to the students.
When all the modules are activated, MLP produces a trajectory within \SI{100}{\ms} and DDPM produces a trajectory within \SI{400}{\ms}.
Compared to the expert that produces a trajectory around \SI{4650}{\ms}, our approach is capable of generating a trajectory with much faster computation time.
Additionally, our trajectory time horizon is roughly \SI{2500}{\ms}, and therefore our approach is capable of replanning trajectories; however, the expert is not capable of replanning trajectories since it takes longer to compute a trajectory than the trajectory time horizon.

\begin{table}[htbp]
    \renewcommand{\arraystretch}{1}
    
    \normalsize
    \centering
    \caption{\centering Test under Out-of-distribution scenarios \textemdash Tighter Constraints \textemdash 100 simulations}
    \label{tab:sim-ood_50_constr}
    \resizebox{\columnwidth}{!}{
        \setlength{\tabcolsep}{1pt}
        \begin{tabular}{>{\centering\arraybackslash}m{0.05\textwidth} >{\centering\arraybackslash}m{0.08\textwidth} >{\centering\arraybackslash}m{0.08\textwidth} >{\centering\arraybackslash}m{0.06\textwidth} >{\centering\arraybackslash}m{0.06\textwidth} >{\centering\arraybackslash}m{0.04\textwidth} >{\centering\arraybackslash}m{0.1\textwidth} >{\centering\arraybackslash}m{0.08\textwidth} >{\centering\arraybackslash}m{0.08\textwidth} >{\centering\arraybackslash}m{0.06\textwidth} }
            \toprule
            & \multicolumn{5}{c}{\textbf{Framework}} & \multicolumn{1}{c}{\textbf{Performance}} & \multicolumn{2}{c}{\textbf{Safety}} & \multirow{2}[10]{*}{\makecell{\textbf{Comp.} \\ \textbf{[ms]}}} \\
            \cmidrule(lr){2-6}
            \cmidrule(lr){7-7}
            \cmidrule(lr){8-9}
            & \# of rollouts & Term. condi. & Colls. avoids. & Guided $t_f$ & QP & Cost & Colls. [\%] & Dyn. violation [\%] & \\
            \midrule
            Expert & - & - & - & - & - & 136.8 & 0.0 & 0.0 & 4649 \\
            \midrule
            \multirow{5}[0]{*}{MLP} & \multirow{5}[0]{*}{-} & \NoRed{} & \NoRed{} & \NoRed{} & \NoRed{} & 23.8 & 0.0 & \textcolor{red}{100.0} & 0.28 \\
                                    & & \NoRed{} & \NoRed{} & \NoRed{} & \YesGreen{} & 3777.3 & \textcolor{red}{52.0} & 0.0 & 44 \\
                                    & & \NoRed{} & \NoRed{} & \YesGreen{} & \YesGreen{} & 2640.3 & 0.0 & 0.0 & 93 \\
                                    & & \NoRed{} & \YesGreen{} & \YesGreen{} & \YesGreen{} & 879.3 & 0.0 & 0.0 & 95 \\
                                    & & \YesGreen{} & \YesGreen{} & \YesGreen{} & \YesGreen{} & 879.4 & 0.0 & 0.0 & 93 \\
            \midrule
            \multirow{5}[0]{*}{DDPM} & \multirow{5}[0]{*}{8} & \NoRed{} & \NoRed{} & \NoRed{} & \NoRed{} & 33.2 & 0.0 & \textcolor{red}{100.0} & 43 \\
                                     &  & \NoRed{} & \NoRed{} & \NoRed{} & \YesGreen{} & 3972.4 & \textcolor{red}{38.0} & 0.0 & 87 \\
                                     &  & \NoRed{} & \NoRed{} & \YesGreen{} & \YesGreen{} & 779.9 & \textcolor{red}{2.0} & 0.0 & 383 \\
                                     &  & \NoRed{} & \YesGreen{} & \YesGreen{} & \YesGreen{} & 528.5 & \textcolor{red}{3.0} & 0.0 & 399 \\
                                     &  & \YesGreen{} & \YesGreen{} & \YesGreen{} & \YesGreen{} & \textcolor{ForestGreen}{\textbf{376.3}} & 0.0 & 0.0 & 399 \\
            \bottomrule
        \end{tabular}}
\end{table}

Next, we test our approach on the scenarios where goal positions are out of the training distribution.
The expert demonstrations, with which the students are trained, are generated with goal positions within \SI{4.0}{\m} from the starting positions, and we test the students on the scenarios where goal positions are \SI{6.0}{\m} away from the starting positions.
Table~\ref{tab:sim-ood_position} shows the benchmarking results for the OOD scenario with different goal positions.
First we tested both MLP and DDPM without any additional constraints, and the results show that both MLP and DDPM violate the dynamic constraints with low performance.
Then we tested both MLP and DDPM with QP, and the results show that both MLP and DDPM do not violate the dynamic constraints, but the performance is significantly degraded and we observe a high collision rate.
Note that even without collision avoidance, the collision rate was \SI{0.0}{\%}, and therefore we did not activate the collision avoidance module for this scenario.
To mitigate the performance degradation, we tested both MLP and DDPM with QP after the denoising steps, and terminal conditioning and guided $t_f$ in the denoising loops.
The number under the columns \textit{Terminal condi.} and \textit{Guided $t_f$} indicates the denoising step when the modules are first introduced.
For instance, the number 4 indicates that the modules are first introduced at $t=4$ denoising step, which is the 2nd denoising step since the first denoising step is at $t=5$.
The results show that DDPM has much better performance than MLP, and as we introduce these two modules earlier in the denoising steps, the performance is improved, showing the modules are more effective when introduced earlier.
Note that when the modules are introduced at $t=5$, the performance is degraded, and this is because in the earlier steps, the trajectories are still quite noisy and the surrogate optimization modules are not effective.

\begin{table}[htbp]
    \renewcommand{\arraystretch}{1}
    \centering
    \caption{\centering Test under Out-of-distribution Scenarios \textemdash Different Goal Positions \textemdash 100 simulations}
    \label{tab:sim-ood_position}
    \setlength{\tabcolsep}{1pt}
    \resizebox{\columnwidth}{!}{
        \setlength{\tabcolsep}{1pt}
        \begin{tabular}{>{\centering\arraybackslash}m{0.05\textwidth} >{\centering\arraybackslash}m{0.08\textwidth} >{\centering\arraybackslash}m{0.04\textwidth} >{\centering\arraybackslash}m{0.06\textwidth} >{\centering\arraybackslash}m{0.04\textwidth} >{\centering\arraybackslash}m{0.1\textwidth} >{\centering\arraybackslash}m{0.08\textwidth} >{\centering\arraybackslash}m{0.08\textwidth} >{\centering\arraybackslash}m{0.06\textwidth} }
            \toprule
            & \multicolumn{4}{c}{\textbf{Framework}} & \multicolumn{1}{c}{\textbf{Performance}} & \multicolumn{2}{c}{\textbf{Safety}} & \multirow{2}[10]{*}{\makecell{\textbf{Comp.} \\ \textbf{[ms]}}} \\
            \cmidrule(lr){2-5}
            \cmidrule(lr){6-6}
            \cmidrule(lr){7-8}
            & \# of rollouts & Term. condi. & Guided $t_f$ & QP & Cost & Colls. [\%] & Dyn. violation [\%] & \\
            \midrule
            Expert & - & - & - & - & 25.0 & 0.0 & 0.0 & 4649 \\
            \midrule
            \multirow{2}[0]{*}{MLP} & \multirow{2}[0]{*}{-} & \NoRed{} & \NoRed{} & \NoRed{} & 4308.5 & 0.0 & \textcolor{red}{92.0} & 0.27 \\
                                    &  & \YesGreen{} & \YesGreen{} & \YesGreen{} & 1942.7 & 0.0 & 0.0 & 91 \\
            \midrule
            \multirow{6}[0]{*}{DDPM} & \multirow{6}[0]{*}{8} & \NoRed{}  & \NoRed{} & \NoRed{} & 4204.5 & 0.0 & \textcolor{red}{100.0} & 42 \\
                                     & & \textcolor{ForestGreen}{1} & \textcolor{ForestGreen}{1} & \multirow{5}[0]{*}{\YesGreen{}} & 571.4 & 0.0 & 0.0 & 399 \\
                                     & & \textcolor{ForestGreen}{2} & \textcolor{ForestGreen}{2} & & 532.3 & 0.0 & 0.0 & 445 \\
                                     & & \textcolor{ForestGreen}{3} & \textcolor{ForestGreen}{3} & & 527.2 & 0.0 & 0.0 & 494 \\
                                     & & \textcolor{ForestGreen}{4} & \textcolor{ForestGreen}{4} & & \textcolor{ForestGreen}{\textbf{517.1}} & 0.0 & 0.0 & 527 \\
                                     & & \textcolor{ForestGreen}{5} & \textcolor{ForestGreen}{5}& & 554.6 & 0.0 & 0.0 & 576 \\
            \bottomrule
        \end{tabular}}
\end{table}

\section{DISCUSSION AND CONCLUSION}\label{sec:conclusion}
We have presented \ourmethod{}, an efficient and highly generalizable IL-based trajectory planning procedure for obstacle avoidance. 

\ourmethod{} divided the computationally challenging problem of finding dynamically feasible and collision-free trajectories into two more tractable sub-problems, solved separately by a diffusion policy and a surrogate optimization problem.

\ourmethod{} performed collision-free trajectory planning via a fast diffusion policy, bypassing challenging computational costs in model-based trajectory planners and exploiting the ability of diffusion policies to well capture highly multimodal output distributions.
Then, the trajectories generated by the diffusion policy were refined by a surrogate optimization problem, which facilitate dynamically feasibility and collision avoidance.
After the diffusion model's denoising, the trajectories were further optimized by a QP to ensure dynamic feasibility. 
Our results showed that the diffusion policy generates collision-free dynamically feasible trajectories with low computational cost.
The generalizability of the diffusion policy was also demonstrated in the OOD scenarios where, at deploymnet, the dynamic constraints were tighter than the training scenarios, and the goal positions were different from the training scenarios.
Future work could investigate its feasibility in more complex scenarios, such as multiagent systems.
\section{Appendix}\label{sec:appendix}

\subsection{Dynamic Constraints in Trajectory Planning}\label{subsec:dynamic-constraints}
    

B-spline curves, $C(t)$, are defined as a linear combination of position control points, $\boldsymbol{P}$, with a set of basis functions, $N_i^p(t)$, as:
\begin{equation}
    C(t) = \sum_{i=0}^{n} \boldsymbol{P}_i N_i^p(t)
\end{equation}
where $n$ is the number of control points, $p$ is the degree of the curve, and $N_i^p(t)$ is the $i$th basis function of degree $p$ evaluated at time $t$. The basis functions are defined recursively as follows:
\begin{eqnarray}
    N_i^0(t) &=& \begin{cases}
        1 & \text{if } t_i \leq t < t_{i+1} \\
        0 & \text{otherwise}
    \end{cases} \\
    N_i^p(t) &=& \frac{t - t_i}{t_{i+p} - t_i} N_i^{p-1}(t) + \frac{t_{i+p+1} - t}{t_{i+p+1} - t_{i+1}} N_{i+1}^{p-1}(t) \notag
\end{eqnarray}
Since the derivative of the basis functions are given by:
\begin{equation}
    \frac{d}{dt} N_i^p(t) = \frac{p}{t_{i+p} - t_i} N_i^{p-1}(t) - \frac{p}{t_{i+p+1} - t_{i+1}} N_{i+1}^{p-1}(t)
\end{equation}
The derivative of the position B-spline curve, which is the velocity B-spline curve, is given by:

\begin{equation}
    \begin{aligned}
        \frac{d}{dt} C(t) &= \sum_{i=0}^{n} \left\{\frac{p}{t_{i+p+1} - t_{i+1}} N_{i+1}^{p-1}(t) (\boldsymbol{P}_{i+1} - \boldsymbol{P}_i)\right\} \\
                            &= \sum_{i=0}^{n} N_{i+1}^{p-1}(t) \boldsymbol{V}_i \\
    \end{aligned}
\end{equation}
where we have defined the velocity control points, $\boldsymbol{V}_i$, as:
\begin{equation}\label{eq:velocity-control-points}
    \boldsymbol{V}_i = \frac{p}{t_{i+p+1} - t_{i+1}} (\boldsymbol{P}_{i+1} - \boldsymbol{P}_i)
\end{equation}
Following the same procedure, the acceleration and jerk B-spline curves are given by:
\begin{equation}
    \begin{aligned}
        \frac{d^2}{dt^2} C(t) &= \sum_{i=0}^{n} N_{i+1}^{p-2}(t) \boldsymbol{A}_i \\
        \frac{d^3}{dt^3} C(t) &= \sum_{i=0}^{n} N_{i+1}^{p-3}(t) \boldsymbol{J}_i \\
    \end{aligned}
\end{equation}
where we have defined the acceleration and jerk control points, $\boldsymbol{A}_i$ and $\boldsymbol{J}_i$, which are given in Eqs.~\ref{eq:acceleration-control-points} and \ref{eq:jerk-control-points}.
\begin{equation}\label{eq:acceleration-control-points}
    \begin{aligned}
    \boldsymbol{A}_i &= \frac{(p-1)}{t_{i+p+1} - t_{i+2}} (\boldsymbol{V}_{i+1} - \boldsymbol{V}_{i}) \\
                    &= \frac{p(p-1)}{t_{i+p+1} - t_{i+2}} \{ \frac{1}{t_{i+p+2} - t_{i+2}} \boldsymbol{P}_{i+2} \\
                    &- \frac{t_{i+p+1} - t_{i+1} + t_{i+p+2} - t_{i+2}}{(t_{i+p+2} - t_{i+2})(t_{i+p+1} - t_{i+1})} \boldsymbol{P}_{i+1} \\
                    &+ \frac{1}{t_{i+p+1} - t_{i+1}} \boldsymbol{P}_{i} \}
    \end{aligned}
\end{equation}

\begin{table*}
    \begin{minipage}{\textwidth}
        \begin{equation}\label{eq:jerk-control-points}
            \begin{aligned}
            \boldsymbol{J}_i &= \frac{(p-2)}{t_{i+p+1} - t_{i+3}} (\boldsymbol{A}_{i+1} - \boldsymbol{A}_{i}) \\
                         &= \frac{p(p-1)(p-2)}{(t_{i+p+1} - t_{i+3})(t_{i+p+1}-t_{i+2})} \\
                         &  \{ \frac{t_{i+p+1}-t_{i+2}}{(t_{i+p+2} - t_{i+3})(t_{i+p+3} - t_{i+3})} \boldsymbol{P}_{i+3} \\
                         &- \frac{(t_{i+p+1} - t_{i+2})(t_{i+p+2} - t_{i+2})(t_{i+p+2} - t_{i+2} + t_{i+p+3} - t_{i+3}) + (t_{i+p+2} - t_{i+2})(t_{i+p+2} - t_{i+3})(t_{i+p+3} - t_{i+3})}{(t_{i+p+2} - t_{i+2})^2 (t_{i+p+2} - t_{i+3})(t_{i+p+3} - t_{i+3})} \boldsymbol{P}_{i+2} \\
                         &+ \frac{(t_{i+p+1} - t_{i+1})(t_{i+p+1} - t_{i+2}) + (t_{i+p+2} - t_{i+3})(t_{i+p+1} - t_{i+1} + t_{i+p+2} - t_{i+2})}{(t_{i+p+1} - t_{i+1})(t_{i+p+2} - t_{i+2})(t_{i+p+2} - t_{i+3})} \boldsymbol{P}_{i+1} \\
                         &- \frac{1}{t_{i+p+1} - t_{i+1}} \boldsymbol{P}_{i} \} 
            \end{aligned}
        \end{equation}
    \end{minipage}
\end{table*}


\subsection{dynamic constraints encoding in QP}

The dynamic constraints on the velocity, acceleration, and jerk of the trajectory can be expressed as:

\begin{equation}\label{eq:dynamic-constraints}
    \begin{aligned}
        \boldsymbol{v}_{\text{min}} &\leq \boldsymbol{V}_i \leq \boldsymbol{v}_{\text{max}}  &\quad i = 0, \dots, n-1 \\
        \boldsymbol{a}_{\text{min}} &\leq \boldsymbol{A}_i \leq \boldsymbol{a}_{\text{max}}  &\quad i = 0, \dots, n-2 \\
        \boldsymbol{j}_{\text{min}} &\leq \boldsymbol{J}_i \leq \boldsymbol{j}_{\text{max}}  &\quad i = 0, \dots, n-3 \\
    \end{aligned}
\end{equation}
where $\boldsymbol{v}_{\text{min}}$, $\boldsymbol{v}_{\text{max}}$, $\boldsymbol{a}_{\text{min}}$, $\boldsymbol{a}_{\text{max}}$, $\boldsymbol{j}_{\text{min}}$, and $\boldsymbol{j}_{\text{max}}$ are the minimum and maximum velocity, acceleration, and jerk constraints, respectively.

By Eqs.~\ref{eq:velocity-control-points}, \ref{eq:acceleration-control-points}, and \ref{eq:jerk-control-points}, the constraints proposed in Eq.~\eqref{eq:dynamic-constraints} can be expressed as a linear inequality constraint on the position control points
\begin{equation}\label{eq:dynamic-constraints-QP}
    A_{\text{dyn}} \boldsymbol{P} \leq b_{\text{dyn}} 
\end{equation}
where $A_{\text{dyn}}$ is a matrix that depends on the time intervals, $\boldsymbol{P}$ is the flattened vector of position control points, and $b_{\text{dyn}}$ is the vector of the minimum and maximum velocity, acceleration, and jerk constraints. 


\balance 
\bibliographystyle{IEEEtran}
\bibliography{main}

\newpage
\end{document}